\documentclass[numbered]{trbunofficial}
\usepackage{graphicx}
\usepackage[hidelinks]{hyperref}
\usepackage{pdflscape}
\usepackage{xcolor}
\usepackage{rotating}
\usepackage{multicol}
\usepackage{multirow}
\usepackage{subcaption}
\usepackage{algorithm}
\usepackage{bbding}
\usepackage{caption}
\usepackage{amsmath}
\usepackage{algpseudocode}
\usepackage{subcaption}
\usepackage{pgfplots}
\usepackage{tabularx}
\usepackage{lipsum}
\usepackage{adjustbox}
\usepackage{booktabs}
\usepackage{colortbl}
\definecolor{myorange}{rgb}{0.8, 0.35, 0}
\definecolor{mygreen}{RGB}{34, 139, 34}

\AuthorHeaders{Liu, Saunier, and Bilodeau}
\title{How good are deep learning methods for automated road safety analysis using video data? An experimental study}

\author{%
  \begin{tabular}{ccc}
    Qingwu Liu & Nicolas Saunier & Guillaume-Alexandre Bilodeau \\
    Polytechnique Montreal & Polytechnique Montreal & Polytechnique Montreal \\
    \texttt{qingwu.liu@polymtl.ca} & \texttt{nicolas.saunier@polymtl.ca} & \texttt{gabilodeau@polymtl.ca}
  \end{tabular}
}

\begin{document}
\maketitle

\section{Abstract}

Image-based multi-object detection (MOD) and multi-object tracking (MOT) are advancing at a fast pace. A variety of 2D and 3D MOD and MOT methods have been developed for monocular and stereo cameras. Road safety analysis can benefit from those advancements. As crashes are rare events, surrogate measures of safety (SMoS) have been developed for safety analyses. (Semi-)Automated safety analysis methods extract road user trajectories to compute safety indicators, for example, Time-to-Collision (TTC) and Post-encroachment Time (PET). Inspired by the success of deep learning in MOD and MOT, we investigate three MOT methods, including one based on a stereo-camera, using the annotated KITTI traffic video dataset. Two post-processing steps, IDsplit and SS, are developed to improve the tracking results and investigate the factors influencing the TTC. The experimental results show that, despite some advantages in terms of the numbers of interactions or similarity to the TTC distributions, all the tested methods systematically over-estimate the number of interactions and under-estimate the TTC: they report more interactions and more severe interactions, making the road user interactions appear less safe than they are. Further efforts will be directed towards testing more methods and more data, in particular from roadside sensors, to verify the results and improve the performance.

\hfill\break%
\noindent\textit{Keywords}: Monocular camera, stereo camera, road user detection and tracking, surrogate measures of safety
\newpage

\section{Introduction}
According to the World Health Organization (WHO), road injury ranked 10\textsuperscript{th} among leading causes of death in both upper-middle-income and lower-middle-income countries and seventh in low-income countries in 2019~\cite{WHO}. The impact of road crashes on society is considerable, from death and hospital treatment costs to traffic congestion and infrastructure repairing costs. The traditional approach to road safety relies on the analysis of historical crash data. Since crashes are relatively rare and random, they require long observation periods before any analysis can be carried out and counter-measures implemented~\cite{elvik1988some}. This is why researchers developed proactive methods that do not require to wait for crashes to occur before road safety issues may be diagnosed. The first such kind of methods were the traffic conflict techniques (TCTs) initiated in the late 1950s by researchers at General Motors~\cite{hayward1971near}. These techniques belong to the more general field of surrogate measures of safety (SMoS) that have drawn considerable interest with the development of more automated data collection methods that alleviate some of the shortcomings of the early manual methods~\cite{saunier2010large,mohamed2015behavior}. The most common SMoS are derived from the observations of user interactions and conflicts and the most frequent approach is to characterize the severity of conflicts and count the numbers of the most severe conflicts. The severity is assumed to reflect both the proximity of a conflict to a crash and the severity of the potential crash. Various objective and subjective indicators have been proposed to measure severity, such as Time-to-Collision (TTC)~\cite{hayward1971near} and Post-encroachment Time (PET)~\cite{allen1978analysis}. These indicators require the trajectories of road users, which are most commonly extracted more or less automatically from video recordings. 

Given their low cost and conveniences, cameras have been widely utilized in traffic data collection. Based on the camera configurations, camera setups can be categorized as monocular, stereo or multi-view. For stereo cameras, the disparities between the left and right images can be utilized to generate accurate depth information. Depth plays an important role on accurately measuring the position of the road users and their direction of motion. While monocular camera cannot provide depth information, neural networks can be trained to learn a representation that distils depth directly, which makes it possible to estimate depth from monocular camera without disparity information. For the object detection and tracking tasks, the depth information is implicit and encoded as the 3D coordinates of objects. For example, 3D CenterNet and 3D CenterTrack regard the depth as an output for generating 3D bounding boxes~\cite{zhou2019objects, zhou2020tracking}. In this study, we selected one stereo camera-based method and two monocular camera-based methods to investigate their performance and whether stereo information can help road user detection, tracking and safety analysis.

The main objective of this research is to investigate the usefulness of stereo information and the performance of state-of-the-art deep learning detection and tracking methods to calculate the TTC. Our contributions can be summarized as follow: 
\begin{itemize}
\item We propose a deep learning-based object detection, tracking and safety analysis framework; 
\item We compare three tracking methods, using monocular and stereo image data, to compute a safety indicator, namely the TTC.
\end{itemize}

\section{Literature~Review}
\subsection{Object~Detection}
Before the advent of deep neural networks (DNNs), the object detection algorithms followed a basic architecture: 1) an image as the input is passed into the model; 2) sliding windows with different sizes and ratios generate regions for analysis; 3) hand-crafted features are extracted for each interesting region, such as Histograms of Oriented Gradients (HOG)~\cite{dalal2005histograms}, Haar wavelets~\cite{lienhart2002extended} and Scale-Invariant Feature Transform (SIFT)~\cite{lindeberg2012scale}, 4) object categories are evaluated for each candidate region by a classification algorithm, such as Support Vector Machine (SVM)~\cite{cortes1995support} and Adaboost~\cite{990517}. 

With the increase of computing power and the occurrence of many large image datasets, methods based on hand-crafted features were gradually replaced by DNN-based object detection methods. Convolutional Neural Network (CNN)-based object detection methods can generally be classified into 1) multi-stage detectors, such as Region-based Convolutional Neural Network (R-CNN)~\cite{girshick2014rich}, Fast R-CNN~\cite{girshick2015fast} and Faster R-CNN~\cite{ren2015faster}, and 2) single-stage detectors, such as Single Shot multiBox Detector (SSD)~\cite{liu2016ssd} and the You Only Look Once (YOLO) detector series~\cite{redmon2016you, redmon2017yolo9000}. While these bounding box-based methods provided improvements, they have the drawback of using anchors that limits the number of objects that can be detected in close proximity. In CornerNet~\cite{law2018cornernet}, the authors proposed a keypoints-based method as an alternative way to represent bounding boxes by keypoints for object representation, which showed a better performance and faster speeds. Two object corners were detected for object location. Following the idea of representing object by keypoints, CenterNet~\cite{zhou2019objects} represents each object by a single point, its center point, making CenterNet the state of the art (SOTA) when it was published. 

\subsection{Object~Tracking}
According to the number of objects to be tracked, object tracking can be separated into two tasks: 1) Visual Object Tracking (VOT), which aims to locate only one object in the frames of a video given the ground-truth information in the first frame and 2) Multiple Object Tracking (MOT), which, instead of tracking a single object, detects all objects of interests and maintains their tracked identities~\cite{luo2021multiple} throughout a video. In the following part, we will mainly cover methods for MOT, since several road users are involved when calculating SMoS and no ground truth is available to initiate tracking in large traffic datasets.

Deep learning-based tracking by detection methods generally follow four basic steps: 1) object detection, 2) feature extraction/motion prediction, 3) affinity calculation and 4) data association ~\cite{ciaparrone2020deep}. For step 2), CNNs-based  methods are widely used for the feature extraction step. As one of the first groups who utilized CNN for feature extraction, ~\citep{kim2015multiple} applied a pretrained CNN for extracting features from the detections, ranking best when it was published. Following the same inspiration, a custom residual CNN was used in SORT~\cite{bewley2016simple} and DeepSORT~\cite{wojke2017simple} to extract visual information and a Fast R-CNN based on VGG-16~\cite{simonyan2014very} was utilized for visual features extractions~\cite{hu2018automatic}. Siamese networks~\cite{kim2016similarity} are trained CNN with loss functions and remove the loss layer for extracting features. To predict the motions of objects,~\citep{wang2017online} employed a correlation filter to generate a response map that predicts the new positions of objects in next frame. For step 3), cosine distance between the extracted features is commonly used for the affinity calculation step~\cite{wojke2017simple, lu2017online}. For step 4), apart from the Kalman filter~\cite{welch1995introduction} and its derived methods that treat association as an estimation and prediction problem, the association of objects among consequent frames can be regarded as a graph bipartite matching problem. Hungarian algorithm~\cite{kuhn1955hungarian} is widely utilized to solve such a graph bipartite matching problem. Besides, apart from tracking by detection methods and detection free tracking (DFT), joint object and tracking methods train a single neural network model for both object detection and tracking task. For example, ~\citep{zhou2020tracking} proposed a method named CenterTrack, which, based on the remarkable performance of CenterNet~\cite{zhou2019objects}, achieved the simultaneous detection and tracking with a single CNN-based architecture and has inspired the following researches.
In this study, we investigate the performances of both tracking methods on road safety analysis.

\subsection{Surrogate Measures of~Safety}
\begin{figure}[h]
    \centering
    \scriptsize
    \includegraphics{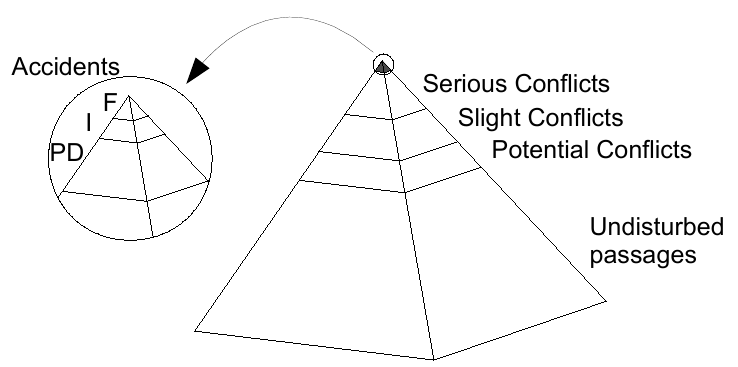}
    \caption{The safety pyramid, adapted from ~\cite{laureshyn2016review}. F refers to the fatal crashes, I to injury crashes and PD to property damage only crashes.}
    \label{fig:safety pyramid}
\end{figure}

Road safety analyses have generally relied and still rely on crash data. Crashes are rare, random, and there is a lack of information in crash reports on the several factors that lead to a crash such as the behaviours of the involved road users. Waiting for crashes to diagnose safety is intrinsically reactive, there is a lot of interest and research in proactive methods, in particular to analyze interactions and near-miss traffic events. Instead of dealing with crashes data, traffic conflict techniques (TCTs) are based on the study of traffic events, whose types are related and represented by the famous safety pyramid, shown in Figure \ref{fig:safety pyramid}. Surrogate measures of safety such as the number of near misses can be derived from the direct observation of traffic events. 

Compared with crash-based methods, traffic event-based methods, including SMoS, show many advantages, such as shorter periods of data collection, smaller collecting costs and richer data. Generally, the most common safety indicators include 1) speed and acceleration, 2) time-to-collision (TTC), which is defined as the time remaining until two road users collide if their movements remain unchanged ~\cite{hayward1972near}, and 3) post-encroachment time (PET), which is defined as the duration between the leaving time of the first road user and the arriving time of the second road user in the crossing zone~\cite{laureshyn2016review, SAUNIER2021662}.

\subsection{Data Extraction Methods for~SMoS} 
At the early stage of traffic conflict techniques (TCTs), manual data collection methods have played the main role in traffic safety researches from the 1960s to the 1990s. Manual data collection works are still employed in some recent studies. For example, in order to validate SMoS, ~\citep{johnsson2021relative} adopted a manual identification method to extract the information of the encounters in six selected intersections. Then the trajectories of each encounter were manually annotated in the T-Analyst software ~\cite{johnsson2020t}. However, it is time-consuming to extract data manually, especially for large/long-time recording data. The demand for more efficient trajectories generation methods pushed the developments of more or less automatic methods to generate trajectories.

In 2012, \citep{saunier2012} released the an open-source ``Traffic Intelligence'' (TI) project, which contains a feature-based tracker for road user detection and tracking, as well as other utilities for trajectory and road safety analysis. Many projects have relied on TI to study the safety of various facilities such as cycle tracks, highway lane markings, roundabouts and automated shuttles. These case studies generally require to further improve the accuracy of the generated trajectories: for example,~\citep{beauchamp2022study} manually verified and corrected the results generated by the TI tracker. 

In the 2010s, several smartphone GPS-based trajectory extracting methods were studied. Johnson et al. developed a smartphone-based system to identify aggressive driving behaviours of drivers in three vehicles ~\cite{johnson2011driving}. ~\citep{stipancic2018vehicle} utilized the GPS-enabled smartphones to collect the traffic data for calculating SMoS. However, those smartphone based methods has a drawback that no enough data will be collected where the sites/locations are difficult to reach.

\section{Methodology}
\subsection{Overview}
\begin{figure}[h]
    \centering
    \includegraphics[width=1\textwidth]{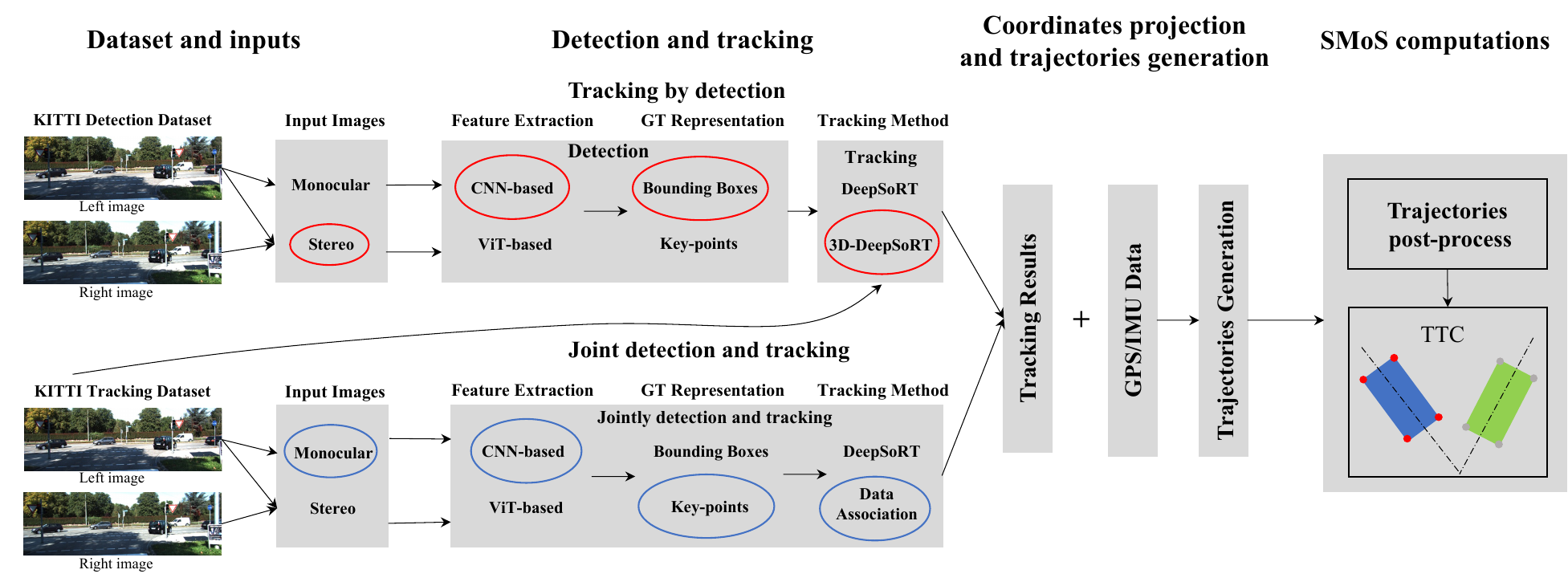}
    \caption{Processing pipeline for deep learning-based road safety analysis. GT represents ground truth, CNN represents convolutional neural network~\cite{girshick2014rich} and ViT  represents vision transformer~\cite{dosovitskiy2020image}. The examples for two object detection and tracking methods, that perform both steps either separately or jointly, are highlighted by the {\color{red} red} ellipses and the {\color{blue} blue} ellipses, respectively. Details of the trajectory post-processing steps are illustrated in Figure~\ref{fig:Post_process}.}
    \label{fig:Framework}
\end{figure}

In this study, we develop a framework for safety analysis using either monocular and stereo cameras. Three tracking methods are tested in the proposed framework. Trajectories are first generated from three deep learning-based object detection and tracking methods that differ in terms of 1)~the type of input, monocular or stereo images, and 2)~the type of tracking method, tracking by detection or joint detection and tracking. Secondly, as opposed to most existing methods, in which the objects are represented by their centroids, the real world coordinates of the road user volumes (3D boxes) are used thanks to the coordinate transferring parameters~\cite{geiger2012we}. The safety indicators are then computed from each vehicle ground level bounding box, also called the bird-eye view (BEV).
The safety analysis framework is shown in Figure \ref{fig:Framework}. It is composed of four parts: 
\begin{enumerate}
    \item dataset and inputs,
    \item detection and tracking, 
    \item coordinates projection and trajectories generation, and 
    \item safety indicator computation. 
\end{enumerate}

\begin{figure}[h]
    \centering
    \includegraphics[width=0.8\textwidth]{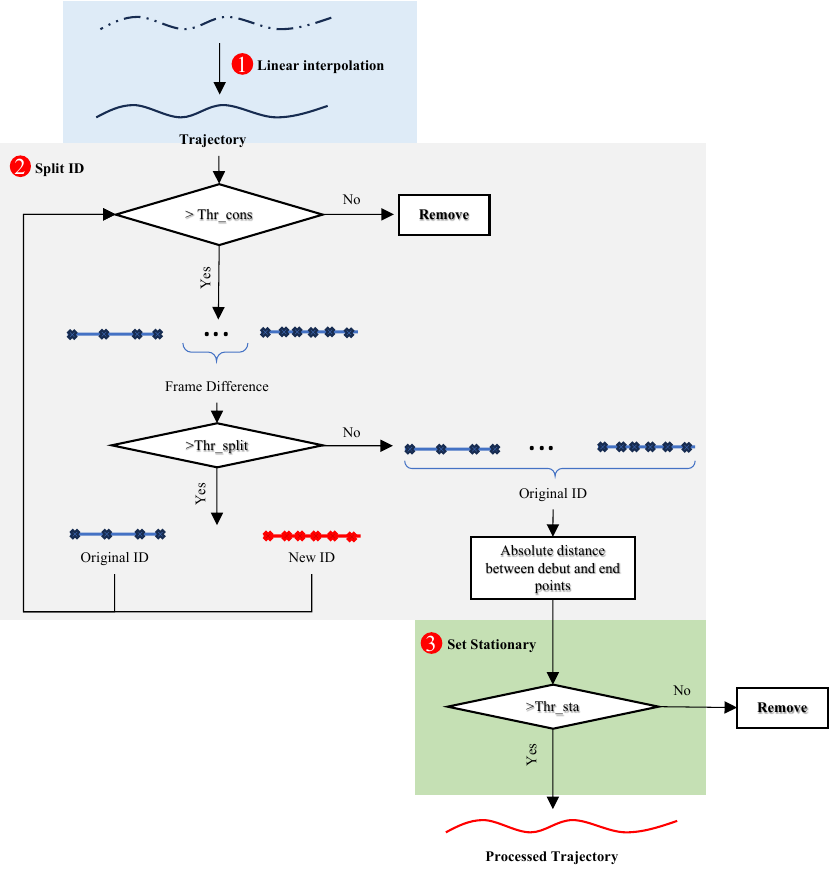}
    \caption{Post-processing steps for road safety analysis.} 
    \label{fig:Post_process}
\end{figure}

\subsection{Object Detection and~Tracking}
Among the three object detection and tracking methods, the first two belong to the category of tracking by detection, and the third performs joint detection and tracking. They are summarized in Table \ref{table:TrackingMethods} and described below:

\begin{table}[!htb]
    \centering
    \scriptsize
    \resizebox{0.9\textwidth}{!}{
    \begin{tabular}{cc|c|c|c|c}
        \hline
        \multicolumn{2}{c|}{\textbf{Detection and tracking method}} & \textbf{Tracking Category} &
          \multirow{2}{*}{\textbf{Input}} &
          \multirow{2}{*}{\textbf{Feature extractor}} &
          \multirow{2}{*}{\textbf{Object Representation}} \\ \cline{1-2}
        \multicolumn{1}{c|}{\textbf{Detection}}    & \textbf{Tracking}    &                                        &           &           &                \\ \hline \hline
        \multicolumn{1}{c|}{MonoDTR}      & 3D-DeepSORT & Tracking by detection & Monocular & ViT & Bounding Boxes \\ \cline{1-2} \cline{4-6} 
        \multicolumn{1}{c|}{YoloStereo3D} & 3D-DeepSORT &                                        & Stereo    & ResNet & Bounding Boxes \\ \hline
        \multicolumn{2}{c|}{CenterTrack}                & Joint detection and tracking  & Monocular & Stacked hourglass & Keypoints     \\ \hline
    \end{tabular}
    }
    \caption{Details of the three methods tested for generating 3D object detection and tracking results.}
        \label{table:TrackingMethods}
\end{table}

\begin{enumerate}
    \item  \textbf{MonoDTR}: As the first method, we select MonoDTR because it is based on a monocular input and it utilizes a depth-aware transformer to achieve 3D object detection~\cite{huang2022monodtr}. In MonoDTR, a novel depth positional encoding was implemented to deal with the problem of weak depth positional performance in transformers. 3D tracking is performed with a modified version of DeepSORT~\cite{wojke2017simple} by computing 3D intersection over union (IoU) instead of 2D IoU;
    \item \textbf{YoloStereo3D}: As the second method, we choose YoloStereo3D~\cite{liu2021yolostereo3d} for detection as it is a stereo image-based 3D object detection method, which relies on ResNet~\cite{he2016deep} as the backbone and utilizes the ghost module~\cite{han2020ghostnet} to improve the accuracy of 3D object detection. Tracking is also performed with the 3D version of DeepSORT;
    \item \textbf{CenterTrack}: As a third method, we choose 3D CenterTrack~\cite{zhou2019objects} that conducts joint detection and tracking from monocular image inputs and generates 3D tracking results akin to the two previous methods.
\end{enumerate}

The outputs of the tracking module follow the format of KITTI~\cite{geiger2012we} and include the 3D center coordinates, 3D box size and the object rotation angles around the y-axis, which are employed for coordinate projection and trajectory generation.

\subsection{Post-Processing Steps for Safety~Analysis}
In the videos recorded by fixed roadside cameras, a static zone is generally selected as the analysis zone where every pair of coexisting road users is analyzed as an interaction~\cite{mohamed2013motion, mohamed2015behavior, beauchamp2022study}. However, videos sequences in the KITTI dataset were recorded by a vehicle-mounted sensor system~\cite{geiger2012we}, which means that such a fixed analysis zone would contain few interactions. In this study, the same definition of interaction is used, where all pairs of simultaneously tracked road users are analyzed, including with the moving vehicle collecting the data. TTC is computed for each interaction at each instant.  

Most video-based methods used for SMoS represent the road users by their centroids~\cite{saunier2006feature,beauchamp2022study} or polygons based on appearance features. This study takes advantage of the chosen methods where objects are represented by ground-level bounding boxes (BEV). TTC can be computed at each instant and depends on a motion prediction method. The simplest method is constant velocity, where the road users are assumed to continue moving in the same direction at constant speed. More realistic and robust methods have been proposed~\cite{mohamed2013motion} and are used~\cite{beauchamp2022study}, but this study starts with the simple constant velocity method. TTC is the first time instant in the future at which the BEV bounding boxes of the two road users overlap, with a maximum value of 10~s. Although TTC cannot be generally computed at each instant, i.e., the road users may not be currently on a collision course at constant velocity, there can be several TTC measures at different instants for a given interaction. Various statistics can be derived to represent the interaction severity, with the minimum value $TTC_{min}$ being one of the most commonly used~\cite{laureshyn2016review}. 

To improve the performance of some detection and tracking methods and to investigate the factors influencing the TTC, we developed the following post-processing steps for the trajectories: 
\begin{enumerate}
    \item In some complex traffic scenarios, however, the data association step may connect different road users if the features of those road users are found to be similar, even when there are large temporal gaps between those tracked road users, e.g., over 40 frames. This kind of tracking error leads, after the next step if these objects are kept, to the generation of a large amount of interactions, which results in a dramatic increase of the amount of dangerous interactions. We therefore apply a trajectory splitting process to handle this problem and compare the results with and without the track splitting process. The track splitting process, called IDsplit in the results, is made of two steps: 1) given a splitting threshold $Thr_{split}$, trajectories with a temporal gap larger than $Thr_{split}$ are split at each gap ($Thr_{split}=10$ in this study). 2) Given a threshold $Thr_{cons}$ on the number of consecutive frames, the new trajectories are removed if they have fewer frames than $Thr_{cons}$, which is selected as 3 in this study. This situation occurs only for tracking-by-detection methods. Since CenterTrack does not connect objects with such long gaps, this step is not applied to CenterTrack. 
    \item In cases of missed detections, for example, caused by occlusions, the road user coordinates may be missing for several frames. For the analysis, they are imputed through linear interpolation. 
    \item The implemented tracking methods detect and track stationary road users as well. Yet, even in the ground truth, the coordinates of these road users will not be exactly constant, resulting in apparent slight random motions and more interactions with measurable TTCs. To explore the impact of these slight position fluctuations on the safety indicators, the last post-processing step, called SS in the results, consists in setting the positions and velocities to the mean position and zero, respectively, if the absolute distance between the debut and end of the $x$ and $y$ coordinates is smaller than $Thr_{sta}$, set to 2.0~m in this study. The interactions are put in two categories, whether the interaction involves a stationary road user, for further analysis.  
\end{enumerate}

\section{DATASETS AND EVALUATION~METRICS}
\subsection{Datasets and Evaluation Metrics}
The following datasets are used in the experiments:
\begin{description}
    \item[KITTI detection]: The KITTI detection benchmark consists of 7,481 training frames and 7,518 test frames. Those sequences are collected by a camera mounted on a car driving through different traffic scenarios. Following the splitting procedure in ~\cite{chen2020dsgn}, the available training set is split into 3,712 training frames and 3,769 validation frames in order to fine-tune YoloStereo3D and MonoDTR to track cars, pedestrians and cyclists instead of only cars and pedestrians~\cite{chen20153d}.
    \item[KITTI tracking]: The KITTI tracking benchmark consists of 21 training sequences and 29 test sequences ~\cite{geiger2012we}. Those sequences are collected by a camera mounted on a car moving through traffic similarly to the KITTI detection dataset. The KITTI tracking ground truth consists of 2D information in pixel coordinates, 3D information in camera coordinates, including height, width, length and the coordinates of the bottom centers of the 3D bounding boxes for cars, pedestrians, and cyclists. Videos are captured with a rate of 10 frames per second (fps). 
\end{description}
In addition to the analysis of safety indicators, we also compute their detection and tracking performance to better understand how it relates to road safety indicators. The object detection performance of YoloStereo3D and MonoDTR is measured in terms of mean Average Precision (mAP) on the validation dataset. The tracking performance is measured using the standard CLEAR MOT metrics~\cite{ristani2016performance}. The main metrics, MOTA and MOTP, are computed using the numbers of false negative $FN$, false positives $FP$, identity switches $IDs$ and ground truth objects $GT$. They rely on the matching of the ground truth objects to the output of each tracking method, generally using the Hungarian algorithm on a cost matrix based on the distance between object centers in image space. We also use the mostly tracked $MT$, mostly lost $ML$, $IDF1$, $MODA$ and $MODP$ as metrics~\cite{ristani2016performance}. MOTA and MOTP are computed using the following equations:

\begin{equation}
    MOTA=1-\frac{FN+ FP+ IDs}{GT} 
\end{equation}

\begin{equation}
    MOTP=\frac{\sum_{t, i} d_{t, i}}{\sum_t c_t},
\end{equation}

\noindent where $d_{t,i}$ is the distance between the position of a ground-truth object $i$ and its matched predicted object. $c_{t}$ is the number of matches made between ground truth and the detection output in frame $t$. $IDF_{1}$, $MODA$ and $MODP$ are computed as: 

\begin{equation}
    IDF_{1}=\frac{2TP}{2TP+FP+FN},
\end{equation}
\begin{equation}
    MODA = \frac{|TP| - |FP|}{|TP|+ |FN|}
\end{equation}
\begin{equation}
        MODP =\sum_{t=1}^{N} \frac{ \sum_{i}^{c_{t}}IoU_{i}^{t}}{c_{t}}
\end{equation}
\noindent where $N$ represents the number of frames and 
$IoU_{i}^{t}$ the intersection over union ($IoU$) of the ground truth object $i$ with its matched predicted object at frame $t$.  
\section{Results and~Discussions}

\subsection{Detection and Tracking Performance}
YoloStereo3D and MonoDTR are fine-tuned for three categories, car, pedestrian and cyclist. For training, we used a computer with a 32GB  NVIDIA Tesla-V100 GPU and dual Intel(R) Xeon(R) Silver 4116@2.10GHz CPUs. CenterTrack was retrained using a computer with four 32GB V100 Volta GPUs and dual Intel Silver 4216 Cascade Lake@2.1GHz CPUs. The main training parameters, such as epoch and learning rate, have the same values as the original methods. 
The results are generated on the same computers as for the training process.

The object detection performance of YoloStereo3D and MonoDTR is presented in Table~\ref{tab:detection results}. YoloStereo3D seems superior to MonoDTR as it is close to MonoDTR for cars (except for the hard cases) and well above MonoDTR for the pedestrian and cyclist classes.  

\begin{table}[h]
    \centering
    \resizebox{0.9\textwidth}{!}{
    \begin{tabular}{c|ccc|ccc|ccc}
        \toprule
        \textbf{Classification} & \multicolumn{3}{c|}{\textbf{Car}}                       & \multicolumn{3}{c|}{\textbf{Pedestrian}}               & \multicolumn{3}{c}{\textbf{Cyclist}}                   \\ \midrule
        \textbf{Method} &
          \multicolumn{1}{c|}{Easy} &
          \multicolumn{1}{c|}{Moderate} &
          Hard &
          \multicolumn{1}{c|}{Easy} &
          \multicolumn{1}{c|}{Moderate} &
          Hard &
          \multicolumn{1}{c|}{Easy} &
          \multicolumn{1}{c|}{Moderate} &
          Hard \\ \midrule
        MonoDTR        & \textbf{99.3} & \textbf{90.84} & \textbf{80.9} & 30.8          & 22.6          & 19.2          & 22.9          & 17.1          & 14.3          \\
        YoloStereo3D   & 97.1          & 86.7           & 69.2          & \textbf{81.1} & \textbf{54.7} & \textbf{41.4} & \textbf{70.4} & \textbf{45.4} & \textbf{34.6} \\ \bottomrule
        \end{tabular}
        }
    \caption{Detection performance (mAP) on the training dataset for the three classes (easy, moderate and hard represent the difficulty of detection with respect to the severity of occlusions).}
    \label{tab:detection results}
\end{table}

For the two object detection methods, MonoDTR and YoloStereo3D, the tracking results are generated by 3D-DeepSORT with the detection results as inputs. For CenterTrack, the training dataset, containing 21 sequences, is split repeatedly into training and validation set using cross validation. Specifically, the training dataset is divided into eighteen sequences as training set and three sequences as validation set, which is repeated seven times. The tracking performance of three methods on the car, pedestrian and cyclist object categories are shown in Table~\ref{tab:trackingresult}. CenterTrack performs the best based on MOTA by a wide margin, and on most other metrics. 

\begin{table}[h]
    \begin{subtable}[h]{\textwidth}
        \centering
        \resizebox{\textwidth}{!}{
        \begin{tabular}{c c c c c c c c c c c c}
            \toprule & 
            MOTA$\uparrow$  & 
            MOTP$\uparrow$  &
            MODA$\uparrow$  & 
            MODP$\uparrow$  &
            MT$\uparrow$  &
            ML$\downarrow$  & 
            FP$\downarrow$  &
            FN$\downarrow$  &
            IDF1$\uparrow$  & 
            IDSW$\downarrow$  &
            FRAG$\downarrow$  \\
            \midrule
            $\mathrm{MonoDTR}$ & 50.0 &	88.2 &	48.8 &	90.5 &	22.1 & 20.4 &  11.5 & 39.7 &	70.5 &	1042 &	1423 \\
            $\mathrm{YoloStereo3D}$ & 45.7 & 	\textbf{88.8} & 49.3  & 	\textbf{91.2} & 	22.6 & 	27.5 & 	6.96 & 	43.7  & 69.1 & 	963 & 	1199 \\
            $\mathrm{CenterTrack}$ & \textbf{84.1} &	88.4 & 	\textbf{85.3} &	90.4 &	\textbf{85.2} &	\textbf{0.03} &	\textbf{4.28} &	\textbf{10.3} &	\textbf{93.1} & 	\textbf{349} &	\textbf{514}\\
            \bottomrule
        \end{tabular}}
        \caption{Tracking performance for cars.}
        \label{tab:trackingresult_car}
    \end{subtable}
    \\
    \begin{subtable}[h]{\textwidth}
        \centering
        \resizebox{\textwidth}{!}{
        \begin{tabular}{c c c c c c c c c c c c}
        \toprule & 
        MOTA$\uparrow$  & 
        MOTP$\uparrow$  &
        MODA$\uparrow$  & 
        MODP$\uparrow$  &
        MT$\uparrow$  &
        ML$\downarrow$  & 
        FP$\downarrow$  &
        FN$\downarrow$  &
        IDF1$\uparrow$  & 
        IDSW$\downarrow$  &
        FRAG$\downarrow$  \\
        \midrule
        $\mathrm{MonoDTR}$ & 22.0 & 68.9 & 32.4 & 92.7 & 14.4 & 27.0 & 17.2 & 50.4 & 59.6 & 1161 & 1505 \\
        $\mathrm{YoloStereo3D}$ & 34.6 & 69.7 & 42.6 & 92.9 & 14.4 & 29.9 & 5.91 & 51.5 & 63.0 & 890 & 1229 \\
        $\mathrm{CenterTrack}$ & \textbf{66.2} & \textbf{77.3} & \textbf{68.2}  & \textbf{93.4}  & \textbf{53.9} & \textbf{7.8} & \textbf{3.9} & \textbf{27.8} & \textbf{82.1} & \textbf{215} & \textbf{530} \\
        \bottomrule
    \end{tabular}}
        \caption{Tracking performance for pedestrians.}
        \label{tab:trackingresult_pedestrian}
     \end{subtable}
     \\
     \begin{subtable}[h]{\textwidth}
        \centering
        \resizebox{\textwidth}{!}{
        \begin{tabular}{c c c c c c c c c c c c}
            \toprule & 
            MOTA$\uparrow$  & 
            MOTP$\uparrow$  &
            MODA$\uparrow$  & 
            MODP$\uparrow$  &
            MT$\uparrow$  &
            ML$\downarrow$  & 
            FP$\downarrow$  &
            FN$\downarrow$  &
            IDF1$\uparrow$  & 
            IDSW$\downarrow$  &
            FRAG$\downarrow$  \\
            \midrule
            $\mathrm{MonoDTR}$ & 7.69  &	82.4  &	11.0  &	98.2 &	13.5 &	43.2 &   30.8 &	58.1 &	48.5 &	63 &	99 \\
            $\mathrm{YoloStereo3D}$ & 40.6 & 	82.3 & 	42.5 & 	\textbf{98.4} & 24.3 & 	27.0 & 	\textbf{7.2} & 	50.3  & 63.4 & 	76 & 86 \\
            $\mathrm{CenterTrack}$ & \textbf{68.4} &	\textbf{82.6} & 	\textbf{69.7} &	97.4 &	\textbf{75.7} &	\textbf{0} & 17.1 &	\textbf{13.3} &	\textbf{85.2} & 	\textbf{22} &	\textbf{51}\\
            \bottomrule
        \end{tabular}}
        \caption{Tracking performance for cyclists.}
        \label{tab:trackingresult_cyclist}
     \end{subtable}
     \caption{Tracking performance results for MonoDTR (MonoDTR+3D-DeepSORT), YoloStereo3D (YoloStereo3D+3D-DeepSORT) and CenterTrack (3D CenterTrack) on cars, pedestrians and cyclists.}
     \label{tab:trackingresult}
\end{table}

\subsection{Safety Analysis}
\subsubsection{Number of Interactions}
This section presents the results on the number of interactions where TTC can be computed. Two sequences are removed from the 21 training sequences of the KITTI tracking dataset: sequence 12, where no $TTC_{min}$ are below 10~s for YoloStereo3D and the ground truth, and sequence 18, where no $TTC_{min}$ are below 10~s for YoloStereo3D after the post-processing steps. We present detailed results for eight sequences (1, 3, 5, 7, 8, 17, 19 and 20) and the total of 19 sequences in Table~\ref{tab:NumbersInteractionsTTC}. Sequences 1 and 7 were recorded in a scenario with many parked cars. Notably, there are five curved roads in sequence 7.  Sequences 3 and 5 were recorded in simpler scenarios with few stationary road users. Sequence 8 was in a scenario with 368 frames, with both moving and stationary road users. Sequence 17 is a scenario with pedestrians and cyclists crossing the road. Sequence 19 is a complex scenario where only pedestrians and cyclists are moving slowly in a crowded street. Sequence 20 was recorded 
with many vehicles moving slowly and stopping in a traffic congestion scenario.

YoloStereo3D performs best with the minimal numbers of interactions with $TTC_{min}$ below 10 and 1.5~s among the three selected methods, although it still generates a considerable amount of interactions compared to the ground truth, especially the most severe with $TTC_{min}$ below 1.5~s (2564 compared to 164 and 966 compared to 80 without and with the post-processing steps, respectively). With both post-processing steps, both YoloStereo3D and CenterTrack perform the best in three sequences (sequence 17, 19 and 20 for YoloStereo3D and sequence 1, 7 and 8 for CenterTrack) below 1.5~s, respectively. Both post-processing steps result in remarkable decreases compared to the initial methods, as presented in Table \ref{tab:TTCReducingPercentages}. Notably, the SS step has a huge influence on the numbers, even for the ground truth, resulting in reductions of the numbers of interactions with $TTC_{min}$ below 10 and 1.5~s of 55.02~\% and 51.21~\%, respectively. 

\subsubsection{TTC Values}
\begin{figure}[htb!]
    \centering
        \includegraphics[width=0.5\textwidth]{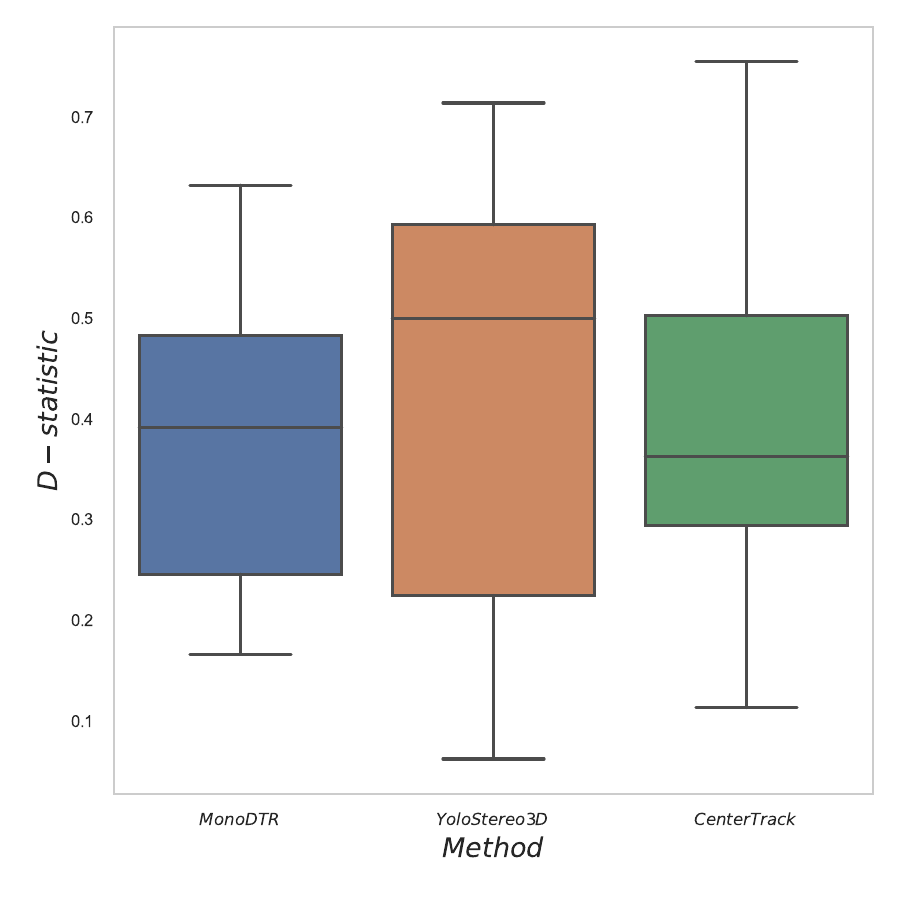}
        \caption{Boxplots of the D-statistics for each method for all the sequences.}
    \label{fig:D-statistic}
\end{figure}

\begin{figure}[htb!]
    \centering
        \includegraphics[width=0.5\textwidth]{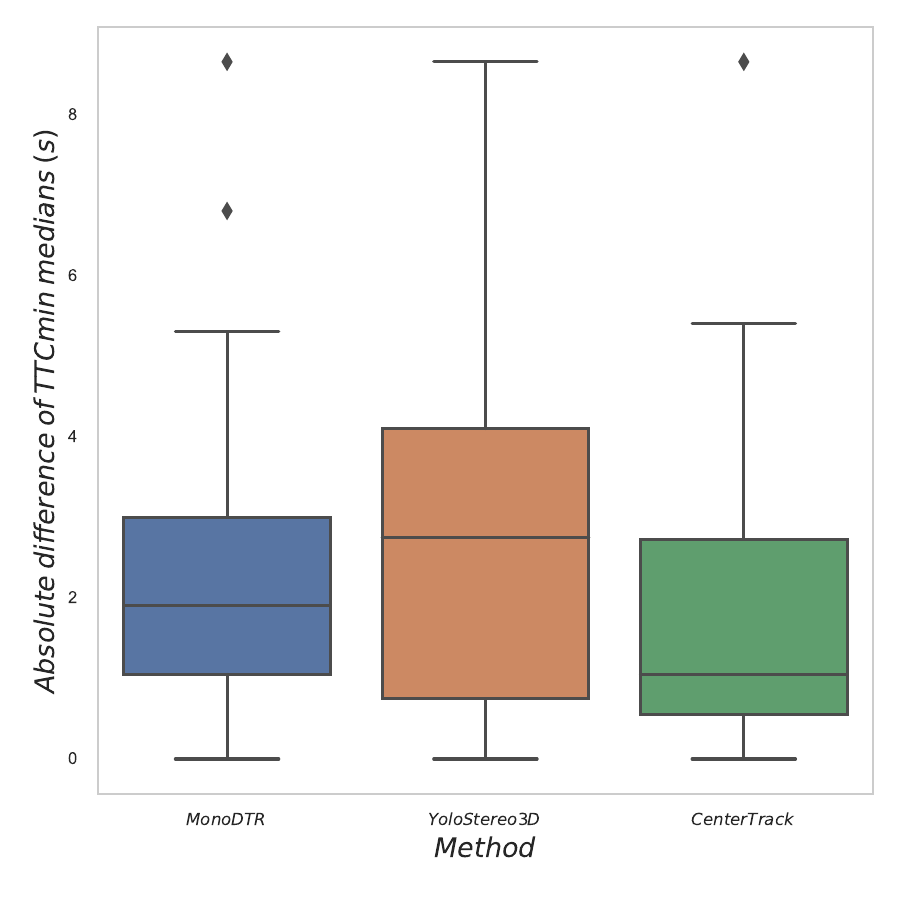}
        \caption{Boxplots of the absolute differences of the $TTC_{min}$ medians between each method and the ground truth for all the sequences.}
    \label{fig:MedianBoxplot}
\end{figure}

\begin{figure}[htb!]
    \centering
        \includegraphics[width=0.6\textwidth]{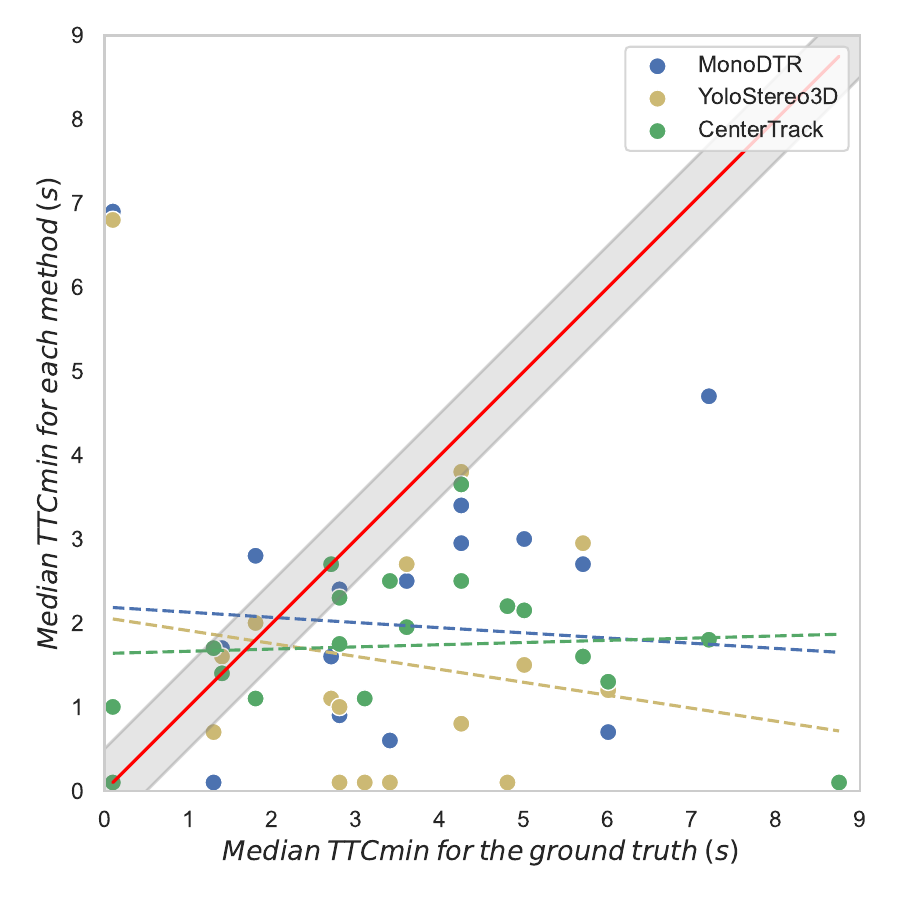}
        \caption{Scatter plot of the median of $TTC_{min}$ for each method as a function of the median of $TTC_{min}$ for the ground truth for all the sequences; the red line is the $y=x$ curve and the gray shaded area represents a range of $\pm$ 0.5~s around the red line.}
    \label{fig:MedianScatterplot}
\end{figure}

The performance of each method should also be measured in terms of the TTC values, not just the number of interactions. The cumulative distribution functions (CDF) of $TTC_{min}$ can be compared between each method and the ground truth through the D-statistic of the Kolmogorov–Smirnov test~\cite{chakravarti1967handbook}, which is computed as the maximum difference between the CDFs. The D-statistic is computed for each method and each sequence and presented as boxplots in Figure~\ref{fig:D-statistic}. Besides, the medians of the $TTC_{min}$ distributions are compared between the output of each method and the ground truth for each sequence in a scatter plot (see Figure~\ref{fig:MedianBoxplot}). The absolute differences between the median of each method and the ground truth for each sequence are depicted in a boxplot (see Figure~\ref{fig:MedianScatterplot}). A shaded region $\pm$0.5~s around the $y=x$ line is added to visualize and quantify the similarities of the medians between the three methods and the ground truth. These results are generated after all the post-processing steps. 

\begin{landscape}
    \begin{table}[h]
        \centering
        \resizebox{1.25\textwidth}{!}{
        \begin{tabular}{c|cc|cc|cc|cc|cc|cc|cc|cc|cc}
            \toprule
            \textbf{Sequence\_number} &
            \multicolumn{2}{c|}{\textbf{0001}} &
            \multicolumn{2}{c|}{\textbf{0003}} &
            \multicolumn{2}{c|}{\textbf{0005}} &
            \multicolumn{2}{c|}{\textbf{0007}} &
            \multicolumn{2}{c|}{\textbf{0008}} &
            \multicolumn{2}{c|}{\textbf{0017}} &
            \multicolumn{2}{c|}{\textbf{0019}} &
            \multicolumn{2}{c|}{\textbf{0020}} &
            \multicolumn{2}{c}{\textbf{Total}}\\ \midrule
    
            \multicolumn{1}{c|}{\textbf{Method}} & 10s  & 1.5s & 10s & 1.5s & 10s & 1.5s & 10s & 1.5s & 10s & 1.5s & 10s & 1.5s & 10s & 1.5s & 10s & 1.5s & 10s & 1.5s \\ \midrule
            MonoDTR  &  \textcolor{blue}{1022}   & \textcolor{blue}{756}  & \textcolor{blue}{36}  & \textcolor{blue}{32}   & \textcolor{blue}{123} & \textcolor{blue}{87}  & \textcolor{blue}{911} & \textcolor{blue}{638}  & \textcolor{blue}{100} & \textcolor{blue}{48}   & \textcolor{blue}{137} & \textcolor{blue}{104}  & \textcolor{blue}{983} & \textcolor{blue}{439}  & \textcolor{blue}{1638} &  \textcolor{blue}{1188} & \textcolor{blue}{6633} & \textcolor{blue}{4087}\\ 
            YoloStereo3D & \textcolor{myorange}{750} & \textcolor{myorange}{588} & \textcolor{myorange}{34} & \textcolor{myorange}{38} & \textcolor{myorange}{128} & \textcolor{myorange}{96} & \textcolor{myorange}{486} & \textcolor{myorange}{288} & \textcolor{myorange}{82} & \textcolor{myorange}{38} & \textcolor{myorange}{63} & \textcolor{myorange}{50} & \textcolor{myorange}{372} & \textcolor{myorange}{124} & \textcolor{myorange}{718} & \textcolor{myorange}{546} & \textcolor{myorange}{4099} & \textcolor{myorange}{2564} \\ 
            CenterTrack  & \textcolor{mygreen}{894} & \textcolor{mygreen}{377} & \textcolor{mygreen}{20} & \textcolor{mygreen}{16} & \textcolor{mygreen}{170} & \textcolor{mygreen}{85} & \textcolor{mygreen}{476} & \textcolor{mygreen}{143} & \textcolor{mygreen}{68} & \textcolor{mygreen}{24} & \textcolor{mygreen}{71} & \textcolor{mygreen}{19} & \textcolor{mygreen}{516} & \textcolor{mygreen}{214} & \textcolor{mygreen}{1140} & \textcolor{mygreen}{668} & \textcolor{mygreen}{6860} & \textcolor{mygreen}{3279} \\
            Ground Truth & \textcolor{red}{157} & \textcolor{red}{30} & \textcolor{red}{10} & \textcolor{red}{4} & \textcolor{red}{26} & \textcolor{red}{8} & \textcolor{red}{114} & \textcolor{red}{10} & \textcolor{red}{14} & \textcolor{red}{0} & \textcolor{red}{1} & \textcolor{red}{0} & \textcolor{red}{88} & \textcolor{red}{9} & \textcolor{red}{138} & \textcolor{red}{0} & \textcolor{red}{1296} & \textcolor{red}{164} \\ \midrule
            MonoDTR+IDsplit & \textcolor{blue}{538} & \textcolor{blue}{322} & \textcolor{blue}{16} & \textcolor{blue}{12} & \textcolor{blue}{133} & \textcolor{blue}{92} & \textcolor{blue}{655} & \textcolor{blue}{339} & \textcolor{blue}{98} & \textcolor{blue}{40} & \textcolor{blue}{90} & \textcolor{blue}{60} & \textcolor{blue}{744} & \textcolor{blue}{213} & \textcolor{blue}{1269} & \textcolor{blue}{712} & \textcolor{blue}{4927} & \textcolor{blue}{2306} \\
    
            YoloStereo3D+IDsplit  & \textcolor{myorange}{470} & \textcolor{myorange}{276} & \textcolor{myorange}{16} & \textcolor{myorange}{12} & \textcolor{myorange}{137} & \textcolor{myorange}{107} & \textcolor{myorange}{342} & \textcolor{myorange}{124} & \textcolor{myorange}{60} & \textcolor{myorange}{30} & \textcolor{myorange}{46} & \textcolor{myorange}{33} & \textcolor{myorange}{353} & \textcolor{myorange}{67} & \textcolor{myorange}{450} & \textcolor{myorange}{218} & \textcolor{myorange}{3243} & \textcolor{myorange}{1446} \\
              \midrule
            \multicolumn{1}{c|}{MonoDTR+IDsplit+SS} & \textcolor{blue}{206} & \textcolor{blue}{182} & \textcolor{blue}{10} & \textcolor{blue}{4} & \textcolor{blue}{26} & \textcolor{blue}{16} & \textcolor{blue}{260} & \textcolor{blue}{128} & \textcolor{blue}{94} & \textcolor{blue}{40} & \textcolor{blue}{59} & \textcolor{blue}{50} & \textcolor{blue}{443} & \textcolor{blue}{115} & \textcolor{blue}{970} & \textcolor{blue}{592} & \textcolor{blue}{2684} & \textcolor{blue}{1430} \\
            YoloStereo3D+IDsplit+SS  & \textcolor{myorange}{192} & \textcolor{myorange}{170} & \textcolor{myorange}{8} & \textcolor{myorange}{8} & \textcolor{myorange}{32} & \textcolor{myorange}{26} & \textcolor{myorange}{112} & \textcolor{myorange}{62} & \textcolor{myorange}{58} & \textcolor{myorange}{30} & \textcolor{myorange}{29} & \textcolor{myorange}{23} & \textcolor{myorange}{205} & \textcolor{myorange}{33} & \textcolor{myorange}{356} & \textcolor{myorange}{200} & \textcolor{myorange}{1685} & \textcolor{myorange}{966} \\
            \multicolumn{1}{c|}{CenterTrack+SS}      & \textcolor{mygreen}{212} & \textcolor{mygreen}{126} & \textcolor{mygreen}{10} & \textcolor{mygreen}{8} & \textcolor{mygreen}{68} & \textcolor{mygreen}{26} & \textcolor{mygreen}{179} & \textcolor{mygreen}{36} & \textcolor{mygreen}{42} & \textcolor{mygreen}{16} & \textcolor{mygreen}{105} & \textcolor{mygreen}{79} & \textcolor{mygreen}{331} & \textcolor{mygreen}{108} & \textcolor{mygreen}{834} & \textcolor{mygreen}{460} & \textcolor{mygreen}{3155} & \textcolor{mygreen}{1496} \\
            GroundTruth+SS  & \textcolor{red}{9} & \textcolor{red}{2} & \textcolor{red}{4} & \textcolor{red}{2} & \textcolor{red}{34} & \textcolor{red}{10} & \textcolor{red}{39} & \textcolor{red}{8} & \textcolor{red}{16} & \textcolor{red}{0} & \textcolor{red}{2} & \textcolor{red}{0} & \textcolor{red}{116} & \textcolor{red}{16} & \textcolor{red}{142} & \textcolor{red}{0} & \textcolor{red}{583} & \textcolor{red}{80} \\
            \bottomrule
        \end{tabular} }
        \caption{Comparisons of numbers of interactions with $TTC_{min}$ below 10 and 1.5~s for each method and post-processing step.}
        \label{tab:NumbersInteractionsTTC}
    \end{table}

    \begin{table}[H]
        \centering
           \begin{tabular}{c|cc|cc}
                \toprule
                \textbf{Post-processing} & \multicolumn{2}{c}{\textbf{IDsplit}} & \multicolumn{2}{|c}{\textbf{IDsplit+SS}} \\ \midrule
                \textbf{Method} & 10s     & 1.5s    & 10s     & 1.5s    \\ \midrule
                MonoDTR         & 25.72\% & 43.58\% & 59.53\% & 65.67\% \\
                YoloStereo3D         & 20.88\% & 43.60\% & 58.89\% & 62.32\% \\
                CenterTrack         &  \XSolidBrush      &   \XSolidBrush      & 54.00\% & 54.38\% \\
                Ground Truth    &     \XSolidBrush    &     \XSolidBrush    & 55.02\% & 51.21\% \\ \bottomrule
            \end{tabular}   
        \caption{Summary of the reductions in the number of interactions with $TTC_{min}$ below 10 and 1.5~s for each method and post-processing step (\XSolidBrush \ indicates a post-processing step is not applied).}
        \label{tab:TTCReducingPercentages}
    \end{table}
\end{landscape}

CenterTrack has the smallest median D-statistic and the smallest inter-quartile range (see Figure \ref{fig:D-statistic}). However, all three methods have high D-statistics, which means that they all exhibit significant disparities with the ground truth. Looking at the boxplots of Figure~\ref{fig:MedianBoxplot}, CenterTrack has the smallest quartiles and MonoDTR has the smallest inter-quartile range but these two methods also have one and two outliers, respectively. This further shows how large the $TTC_{min}$ differences are for all methods. One can also look at the medians values themselves in Figure~\ref{fig:MedianScatterplot}. First, it is clear that most methods under-estimate the $TTC_{min}$ medians for most sequences, sometimes by very large margins. Only a few points lie close to the equality line ($y=x$): five for CenterTrack, four for YoloStereo3D and two for MonoDTR within 0.5~s from the ground truth. Second, the regression lines for each method are mostly horizontal, which means that all computer vision methods tend to generate similar median $TTC_{min}$, independently from the true severity of the data (which itself varies considerably). 


\begin{figure}[htb!]
    \centering
    \includegraphics[width=0.8\textwidth]{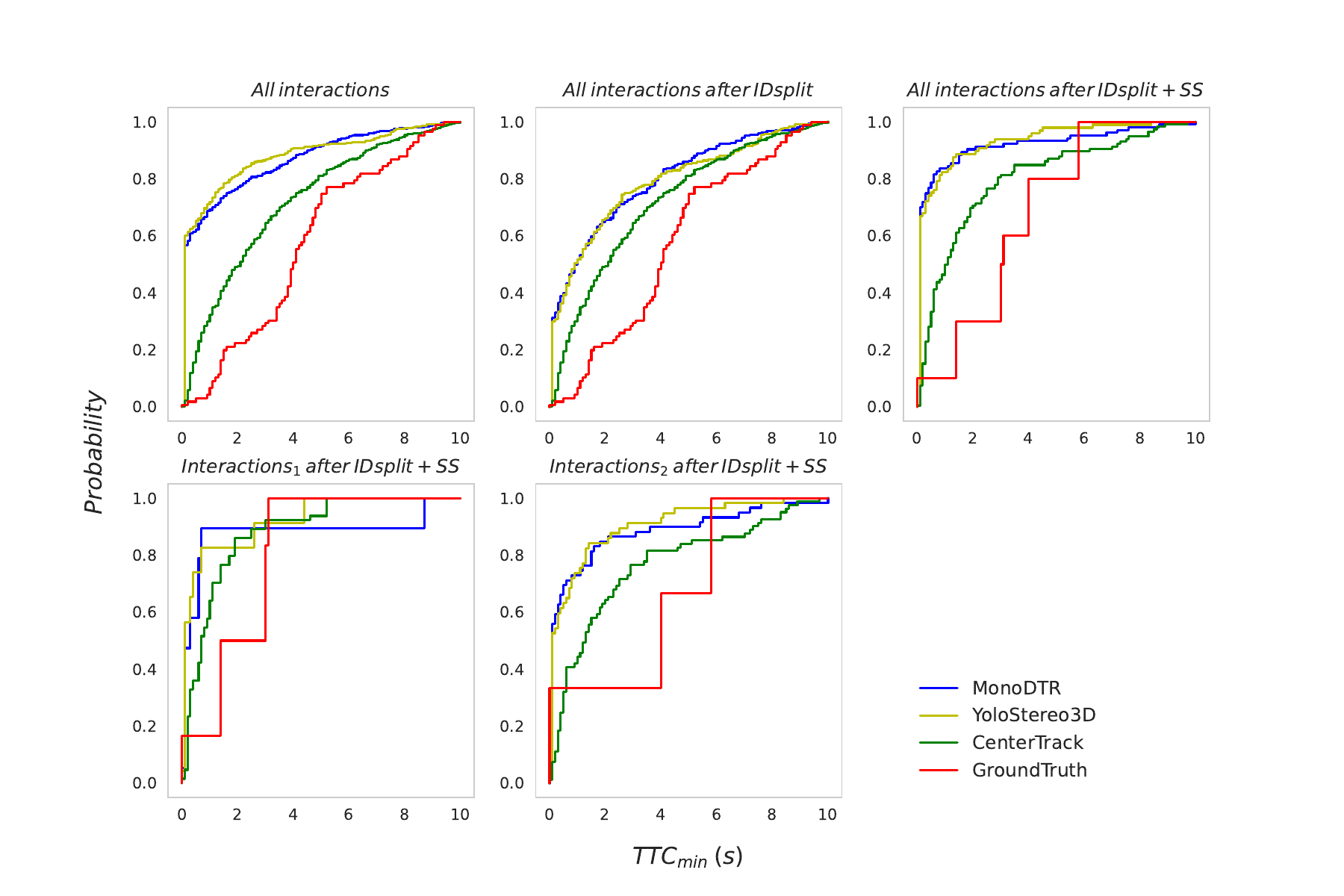}
    \caption{CDFs of $TTC_{min}$ for three methods and ground truth for sequence~1. The top three figures show the CDFs for all the interactions without post-processing, after IDsplit, and after IDsplit+SS. The bottom two figures show the CDFs after IDsplit+SS for the interactions between two moving road users ($Interaction_1$), and the interactions involving a stationary road user ($Interaction_2$), respectively.}
    \label{fig:TTC_seq01}
\end{figure}

\begin{figure}[htb!]
    \centering
    \includegraphics[width=0.8\textwidth]{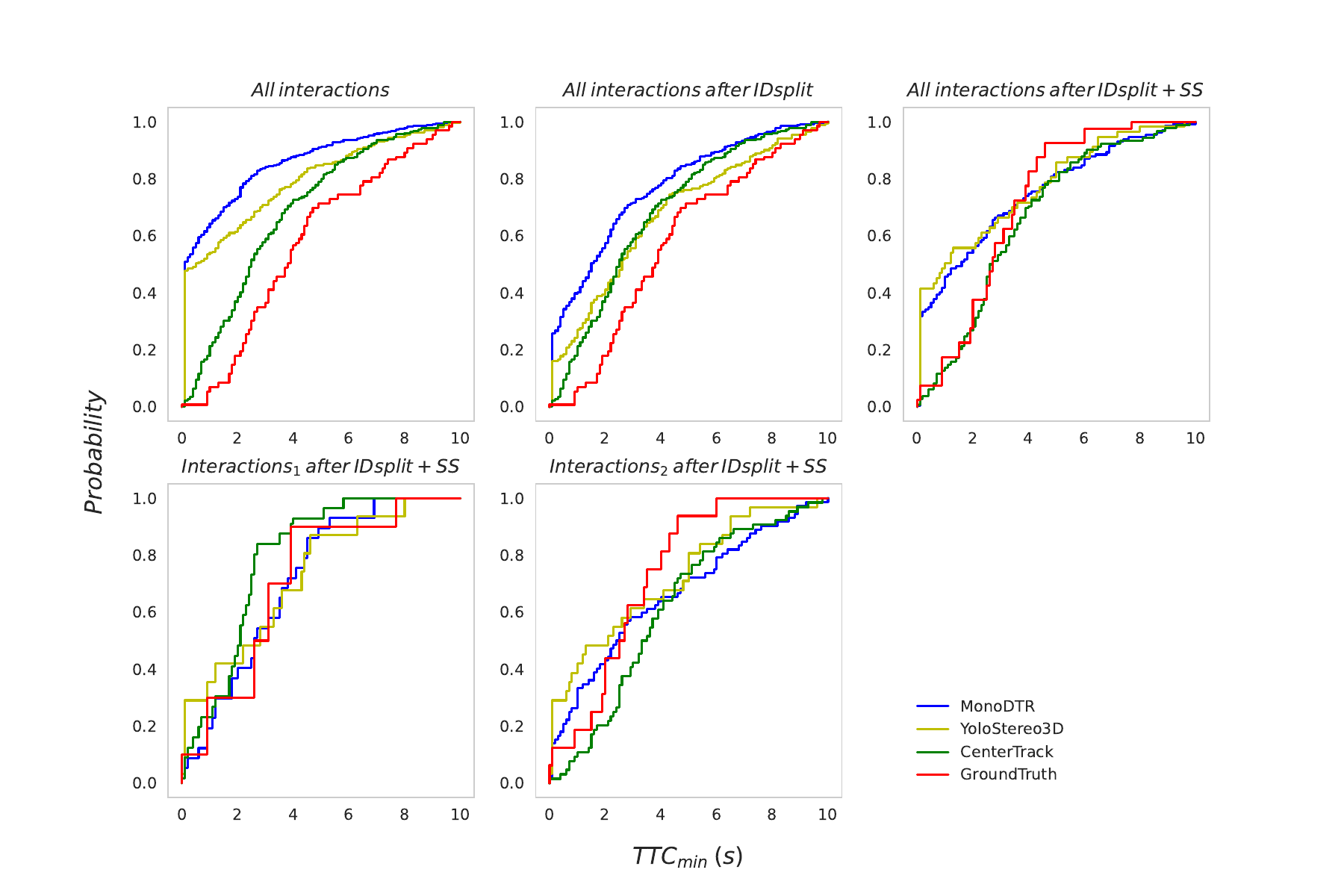}
    \caption{CDFs of $TTC_{min}$ for three methods and ground truth for sequence~7 (see description of Figure \ref{fig:TTC_seq01}).}
    \label{fig:TTC_seq7}
\end{figure}

\begin{figure}[htb!]
    \centering
    \includegraphics[width=0.8\textwidth]{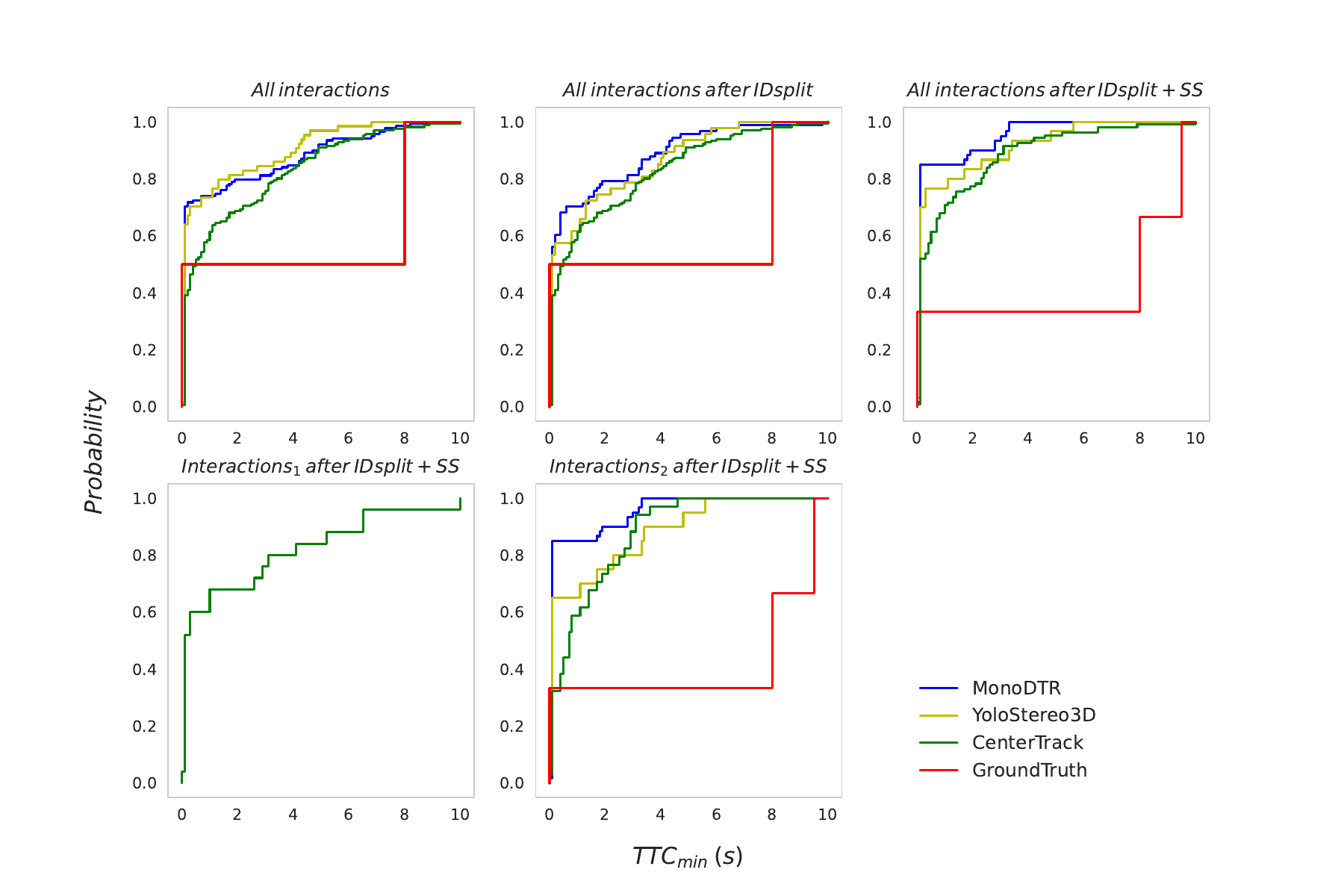}
    \caption{CDFs of $TTC_{min}$ for three methods and ground truth for sequence~17 (see description of Figure \ref{fig:TTC_seq01}). Note there are no $TTC_{min}$ below 10~s in the fourth plot for MonoDTR, YoloStereo3D and GroundTruth.}
    \label{fig:TTC_seq17}. 
\end{figure}

\begin{figure}[htb!]
    \centering
    \includegraphics[width=0.8\textwidth]{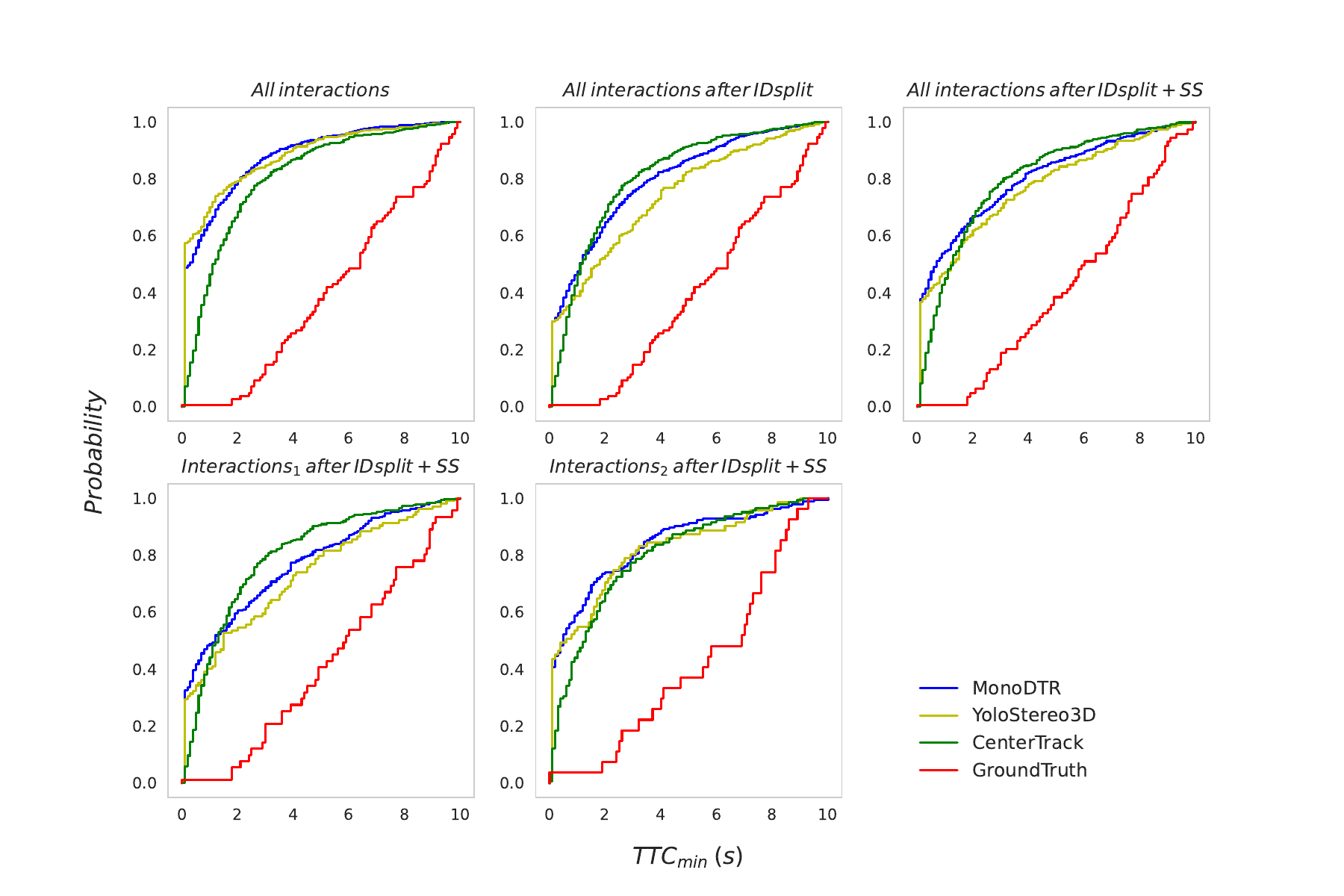}
    \caption{CDFs of $TTC_{min}$ for three methods and ground truth for sequence~20 (see description of Figure \ref{fig:TTC_seq01}).}
    \label{fig:TTC_seq20}
\end{figure}

Finally, let us look at the CDFs of $TTC_{min}$ in detail for sequences 1, 7, 17 and 20 in Figures~\ref{fig:TTC_seq01} to~\ref{fig:TTC_seq20} respectively. For all the four sequences, the IDsplit step reduces the numbers of interactions for MonoDTR, YoloStereo3D and the ground truth. The probabilities of the numbers of interactions below 1.5~s meet considerable reductions for sequences 1, 7 and 20, which indicates that many road users involved in interactions with $TTC_{min}$ below 1.5~s have been incorrectly labeled with the same IDs , compared with those involved in interactions with $TTC_{min}$ between 1.5 and 10~s. However, the proportion of those interactions in sequence 17 only meets a slight reduction, which indicates that the IDsplit step does remove some interactions but its impact is not as pronounced as for sequences 1, 7 and 20. With the SS step,  the numbers of interactions of all three methods and even the ground truth meet large reductions on all four sequences, which indicates that incorrectly measuring the motion of stationary road users has a large influence on the computing of TTC. The slight position fluctuations can lead to numerous inaccurately computed dangerous interactions.

For sequences 1, 7 and 17, the CDFs of the interactions involving a stationary road user ($Interactions_{2}$) are more similar to the CDFs for all interactions after the IDsplit and SS steps, which indicates that the contribution of the interactions involving a stationary road user  ($Interactions_{2}$) overweigh the interactions between two moving road users ($interactions_{1}$). For sequence 20, the CDFs of both $Interactions_{1}$ and $Interactions_{2}$ are similar to the CDFs for all the interactions after the two post-processing steps, which indicates that both types of interactions affect the computing of $TTCmin$ in this traffic congestion scenario. 


\subsection{Discussion}
The results are definitely mixed for the three tested methods and the different sensor types. There is no clear winner. Worse than that, one could see all tested methods as losers given how poorly they reflect the safety as measured by TTC. 

The first part of the results section reported the performance of the methods for the detection and tracking tasks. Comparing these to the safety performance, we see some relationships, as the best tracking method, CenterTrack, is also the slightly better method for TTC. The second best (and superior for detection), YoloStereo3D, is the best in terms of the numbers of interactions. Yet good tracking performance does not translate in accurate TTC values. 

\section{Conclusion}
In this study, we investigated how deep-learning based tracking methods perform for automated road safety analysis. A framework is developed and three different tracking methods, MonoDTR, YoloStereo3D and CenterTrack were selected to compute the road safety indicator TTC on the KITTI tracking dataset. Two post-processing steps are utilized to improve the tracking results and explore the factors that influence the deep-learning based tracking results on TTC. 
Results show that YoloStereo3D outperforms the others with respect to the numbers of interactions with $TTC_{min}$ below 10 and 1.5~s. However, CenterTrack is slightly better in terms of $TTC_{min}$ values as shown by the D-statistic with the ground truth and the $TTC_{min}$ median. Despite these differences and the improvements brought by the proposed post-processing steps, all methods generate many more interactions with much lower $TTC_{min}$ values than in reality. 
The comparisons of CDFs show that interactions involving a stationary road user have a larger impact than interactions with two moving road users on the $TTC_{min}$ distribution, except in a traffic jam scenario where both types of interactions contribute equally.

In conclusion, it is difficult to recommend any of the tested methods. The method based on a stereo-camera shows a slight advantage for some sequences in terms of the number of interactions, but has lower performance than a method based on a monocular camera in terms of the TTC values, and all over-estimate the severity of road user interactions.

Future research will focus on computing and reporting more road safety indicators, such as PET. Efforts will be made to explore and understand the reasons behind the substantial disparities between deep learning methods and the ground truth to develop improved road user detection and tracking methods for safety analysis. Stereo and monocular cameras will be also tested on videos from fixed roadside sensors to verify the current findings.

\section{Acknowledgements}
The authors gratefully acknowledge financial support from the China Scholarship Council. The authors would also like to thank the Interuniversity Research Centre on Enterprise Networks, Logistics and Transportation (CIRRELT) for the computing servers.
\newpage
\bibliographystyle{trb}
\bibliography{reference}
\end{document}